%% file: neurips_2019.tex
\documentclass{article}

\PassOptionsToPackage{numbers, compress}{natbib}

\usepackage[preprint]{neurips_2019}

\usepackage[utf8]{inputenc} 
\usepackage[T1]{fontenc}    
\usepackage{hyperref}       
\usepackage{url}            
\usepackage{booktabs}       
\usepackage{amsfonts}       
\usepackage{nicefrac}       
\usepackage{microtype}      
\usepackage{grffile}
\usepackage{amsmath}
\usepackage{amsthm}
\usepackage{bm}
\usepackage{mathrsfs}
\usepackage{amssymb}
\usepackage{graphicx}
\usepackage{url}
\usepackage{caption}
\usepackage{subcaption}
\usepackage{array}
\usepackage{multirow}
\usepackage{diagbox}
\usepackage{rotating}

  \usepackage{pgfplots}
  \usetikzlibrary{plotmarks}
  \usetikzlibrary{arrows.meta}
  \usepgfplotslibrary{patchplots}
  
\pgfplotsset{
  compat=newest,
  plot coordinates/math parser=false,
  tick label style={font=\scriptsize, /pgf/number format/fixed},
  label style={font=\small},
  every axis/.append style={
    tick align=outside,
    clip mode=individual,
    scaled ticks=false,
    thick,
    tick style={semithick, black}
  }
}

\pgfkeys{/pgf/number format/.cd, set thousands separator={\,}}
\usetikzlibrary{calc}
\usepgfplotslibrary{external}
\pgfplotsset{plot coordinates/math parser=false}
\newlength\figureheight
\newlength\figurewidth
  
\newlength\figH
\newlength\figW
  
\usepackage{array}
\newcommand{\PreserveBackslash}[1]{\let\temp=\\#1\let\\=\temp}
\newcolumntype{C}[1]{>{\PreserveBackslash\centering}p{#1}}
\newcolumntype{R}[1]{>{\PreserveBackslash\raggedleft}p{#1}}
\newcolumntype{L}[1]{>{\PreserveBackslash\raggedright}p{#1}}

\newcommand{\mc}[1]{\mathcal{#1}}
\newcommand{\data}{\mc{D}}
\newcommand{\model}{\mc{M}}
\newcommand{\given}{\mid}
\newcommand{\E}{\mathbb{E}}

\newcommand{\mat}[1]{\bm{\mathrm{#1}}}
\renewcommand{\vec}[1]{\bm{\mathrm{#1}}}
\newcommand{\acro}[1]{\textsc{\MakeLowercase{#1}}}
\newcommand{\R}{\mathbb{R}}

\DeclareMathOperator*{\argmax}{arg\,max}

\title{Accelerating Psychometric Screening Tests With Bayesian Active Differential Selection}

\author{Trevor J.~Larsen \\
  Department of Computer Science and Engineering\\
  Washington University in St. Louis\\
  St. Louis, MO 63130 \\
  \texttt{trevorlarsen@wustl.edu} \\
  \And
  Gustavo~Malkomes \\
  SigOpt \\
  San Francisco, CA 94108 USA \\
  \texttt{gustavo@sigopt.com} \\
   \And
  Dennis L.~Barbour \\
  Department of Biomedical Engineering \\
  Washington University in St. Louis
  St. Louis, MO 63130 \\
  \texttt{dbarbour@wustl.edu} \\
}

\begin{document}

\maketitle

\begin{abstract}
\input{./content/abstract.tex}
\end{abstract}

\input{./content/intro.tex}
\input{./content/method.tex}
\input{./content/experiments.tex}

\bibliography{neurips_2019}
\bibliographystyle{plainnat}

\appendix
\input{./content/appendix}

\end{document}

%% file: content/abstract.tex

Classical methods for psychometric function estimation either require excessive measurements or produce only a low-resolution approximation of the target psychometric function. In this paper, we propose a novel solution for rapid screening for a change in the psychometric function estimation of a given patient. We use Bayesian active model selection to perform an automated pure-tone audiogram test with the goal of quickly finding if the current audiogram will be different from a previous audiogram. We validate our approach using audiometric data from the National Institute for Occupational Safety and Health (\acro{NIOSH}). Initial results show that with a few tones we can detect if the patient's audiometric function has changed between the two test sessions with high confidence.  

%% file: content/intro.tex

\section{Introduction} 

Machine learning (\acro{ML}) has great potential to improve healthcare, with such applications as personalized medicine and automated robotic surgery  
\cite{KONONENKO200189, Hamburg:2010, Shademan337ra64}. 
In particular, \acro{ML} can be use to aid in the judicious application of
healthcare resources in resource-poor settings. An example of this
is implementing the most appropriate use of expensive or intrusive diagnostic procedures. 

Perceptual testing to diagnose disorders of hearing and vision requires 
many repeated stimulus presentations to patients in order to determine their condition.
By carefully controlling the experiment conditions  (\textit{e.g.} 
the strength, duration, or other characteristics of the stimulus), 
clinicians estimate the subject's perception. 
Precisely, the clinician estimates a psychometric function, an inference model 
mapping features of the physical stimulus to the patient response. 
We will focus on hearing tests, but our methodology generalizes to any psychometric test reflecting perceptual or cognitive phenomena.

Traditional audiometry tests, such as the modified Hughson-Westlake procedure \cite{carhart1959preferred},
are low-resolution and time-consuming. 
Clinicians present a series of tones at various frequencies 
(corresponding to pitch) and intensities (corresponding to loudness) 
to the patients and record their response (\textit{i.e.} whether he/she hears the tone). 
A standard audiometry test requires 15-30 minutes, 
and it only estimates the hearing threshold\,---\,the softest sound level one can hear\,--- at a few discrete frequencies. This labor-intensive approach scales poorly to large populations. 
Crucially, some disorders such as 
noise-induced hearing loss can be entirely preventable with a sensitive, early diagnostic.
Unfortunately, the standard methodology greatly hinders rapid screening at high-resolution. 

Alternative tests have been investigated, and 
special interest has been given to procedures using 
Bayesian active learning \cite{Leek2001, Schlittenlacher:2018}.
In this framework, it is possible to leverage audiologist's expertise
to construct automated audiometry tests.  Active learning strategies for estimating a patient's hearing threshold were studied in \citep{Schlittenlacher:2018, Song:2015, gardner2015psychophysical}.
\citet{Gardner:2015} further applied Bayesian active model selection to rapid screening for noice-induced hearing loss.

In this work, we extend previous approaches on Bayesian active learning for audiometry tests by mathematically 
incorporating prior information about the patients. Our framework enables rapid screening for changes in the patient's clinical condition. 
Given a previous audiometry test, our approach automatically delivers a sequence of 
stimuli to quickly determine whether the patient's hearing thresholds are similar or 
different than a reference exam, such as a previous test in the same patient or a population average. 

%% file: content/method.tex

\section{Bayesian Active Differential Selection}

Consider supervised learning problems defined on an input space
$\mc{X}$ and an output space $\mc{Y}$.  
We are given a set of observations $\data = (\mat{X}, \vec{y})$, where $\mat{X}$ represents the design matrix of independent variables $\mat{x}_i \in \mc{X}$ and $\vec{y}$ the associated vector of dependent variables $y_i =
y(\vec{x}_i) \in \mc{Y}$. 
We assume that these data were generated via a latent function $f\colon \mc{X} \to \R$ with a known observation 
model $p(\vec{y} \given \vec{f})$, where $f_i = f(\vec{x}_i)$. 
In this context the latent function $f$ is the psychometric function, 
and the initial data were obtained during a previous exam in the same patient.

Suppose that, after some undetermined period of time following the first exam, we wish to collect a new set of observations $\data'$ from the same 
phenomenon $f$, \textit{e.g.}, the psychometric function in the same individual. 
Our goal is to perform measurements\,---\,select $\vec{x}^\ast \in \mc{X}$ and observe $y^\ast = y(\vec{x}^\ast)$\,---\,to quickly distinguish whether or not the latent function $f$ has changed. In our medical application, this translates to rapid screening for a different clinical condition or a change in condition.

We begin by modeling this as a two-task \textit{active learning} problem
\footnote{We use active learning in its broader sense of intelligently selecting observations to achieve \textit{any} goal, as opposed to restricting it to \textit{learning} predictors with few training samples}.
We define a new input space by augmenting $\mc{X}$ with a feature representing which task (or test) the data points come from, \textit{i.e.}, $\mc{X'} : \mc{X} \times \mc{T}$, where $\mc{T} = \{1,2\}$. For all prior observations we have $\data = (\big[\mat{X}, \vec{1}\big], \vec{y})$ and each new observation will be from the new task, $\vec{x}^\ast = [\vec{x}^T, 2]^T, \vec{x} \in \mc{X}$. 
Next, we hypothesize that the data can be explained by one of two probabilistic models:  $\model_f$, which assumes that $\data$ and $\data'$ come from the same underlying function $f$; and $\model_g$, which offers a different explanation for the most recent set of observations $\data'$. 
Under these assumptions we are interested in selecting candidate locations $\vec{x}^\ast$ to quickly differentiate these two models. We pursue this goal motivated by ideas from information theory, which were successfully applied in a series of active-learning papers \citep{houlsby2011bayesian,garnett2013active,gardner2015psychophysical,hl2014pes,houlsby2014cold, Gardner:2015}. Specifically, we select $\vec{x}^\ast$ maximizing the mutual information between the observation $y^\ast$ and the unknown models:
\begin{align}
  I(y^\ast; \model \given \vec{x}^\ast, \data \cup \data')
  &=
  \label{bald_mi}
  H[y^\ast \given \vec{x}^\ast, \data \cup \data']
  -
  \E_{\model}\!\bigl[
    H[y^\ast \given \vec{x}^\ast, \data \cup \data', \model]
  \bigr],
\end{align}
where $H$ indicates differential entropy \cite{cavagnaro2010adaptive, Gardner:2015}. 
We use the (more-tractable) formulation of mutual information (\cite{Gardner:2015})
that requires only computing the model-conditional predictive distributions:
  $\bigl\{ p(y^\ast \given \vec{x}^\ast, \data \cup \data', \model_i) \bigr\}$,
the differential entropy of each of these distributions, 
and the differential entropy of the
model-marginal predictive distribution:
\begin{equation}
  \label{model_marginal}
  p(y^\ast \given \vec{x}^\ast, \data \cup \data')
  =
  \sum_j p(y^\ast \given \vec{x}^\ast, \data \cup \data', \model_j)\,p(\model_j \given \data \cup \data')
\end{equation}

\textbf{Bayesian Active Differential Selection for Gaussian processes}.
Following a series of work on active learning for audiometry, 
we use a \acro{GP} to model the psychometric function \cite{Barbour2019, Schlittenlacher:2018, DiLorenzo2017ConjointAE, Song:2017, Song:2015, gardner2015psychophysical, Gardner:2015}. 
A Gaussian Process (\acro{GP}) is completely defined by its
first two moments, a mean function $\mu\colon \mc{X'} \to \R$ and a
covariance or kernel function $K\colon \mc{X'}^2 \to \R$. 
For further details on \acro{GP}s see \citep{RasWil06}.


\textbf{Audiometric function model $\model_f$}.
We use a constant prior mean function $\mu_f(.) = c$ to model a frequency-independent natural threshold. 
While audiograms do not necessary have a constant mean, previous research has shown that a constant mean function is sufficient for modeling audiograms, as the covariance function captures the shape of the psychometric function in the posterior distribution \cite{Barbour2019, Song:2015, gardner2015psychophysical}. 
For the covariance function, we use a linear kernel in intensity and a squared-exponential covariance in frequency as proposed by \cite{gardner2015psychophysical, Gardner:2015}. 
Let $[i, \phi]$ represent a tone stimulus, with $i$ representing its intensity and $\phi$ its frequency, and set $\vec{x} = [i, \phi, t] \in \mc{X'}$, where $t$ is the task-related feature.
The covariance function of our first model is independent of the task and given by:
\begin{equation}
  \label{kernel_model_f}
  K_f\bigl([i,\phi, t], [i',\phi', t']\bigr) =
  K_{[i, \phi]}\bigl([i,\phi], [i',\phi']\bigr) =
  \alpha ii' +
  \beta\exp\bigl(-\tfrac{1}{2\ell^2}\lvert \phi - \phi' \rvert^{2}\bigr),
\end{equation}
where $\alpha, \beta > 0$ weight each component and $\ell > 0$ is a length scale of frequency-dependent random deviations from a constant hearing threshold. 
This kernel enforces the idea that i) hearing is monotonic as a function of intensity, 
and ii) the audiometric function should be continuous and smooth across the frequency domain because nearby locations in the cochlea are physically coupled
\cite{gardner2015psychophysical}.

Our $\model_f$ is a $\acro{GP}$ model, $f = \mc{GP}(\mu_f, K_f)$, with cumulative Gaussian likelihood (probit regression), where $y$ takes a Bernoulli
distribution with probability $\Phi(f(\vec{x}))$, and $\Phi$ is the Gaussian \acro{CDF}.
Computing the exact form of the latent posterior distribution is intractable because of the probit 
likelihood function. Thus, the posterior must be approximated. For this model, we use the \emph{expectation propagation} (\acro{EP}) approximation \cite{minka2001expectation, RasWil06}. 
Additionally, we perform inference by maximizing the hyperparameter posterior, finding
the maximum \emph{a posteriori} (\acro{MAP}) hyperparameters:
\begin{equation}
  \label{map_theta}
  \hat{\theta}
  =
  \argmax_\theta
  \log p(\theta \given \data, \model_f)
  =
  \argmax_\theta
  \log p(\theta \given \model_f) +
  \log(\vec{y} \given \mat{X}, \theta, \model_f).
\end{equation}
Notice that we only perform inference after observing $\data$. During the active-learning procedure for constructing $\data'$, we fix the \acro{MAP} hyperparameters, due to our hypothesis that the new observations $\data'$ should not drastically change our belief about $f$. 

\textbf{Differential model $\model_g$}.
The previous model assumes that both $\data$ and $\data'$ can be explained by the same probabilistic model. Now, we present our differential model, which has a less restrictive assumption. We assume that the first set of observations $\data$ was generated by a latent function $f$ and that the new observations $(\vec{x}^\ast, y^\ast) \in \data'$ are generated by a different latent function $g\colon \mc{X}' \to \R$. Specifically, we use the following kernel to capture the correlation between both tasks:
\begin{equation}
  K_g\bigl([i,\phi, t], [i',\phi', t']\bigr) =
  K_t(t, t') \times K_{[i, \phi]}\bigl([i,\phi], [i',\phi']\bigr)
\end{equation}
where $K_{[i, \phi]}$ is the same as in \eqref{kernel_model_f} and the task or conjoint kernel is defined as:
\[
  K_t(t,t')
  =
  \begin{cases}
    1 & t = t' \\
    \rho_t & t \neq t'.
  \end{cases}
\]
The new parameter $\rho_t$ can be interpreted as the correlation between tasks and is referred to as the task or conjoint correlation.
We learn the hyperparameters of this model for both mean and covariance functions.
Hyperparameters associated exclusively to the first task are fixed, similarly to previous model. For the input locations related to the second task $t=2$, we optimize 
$\theta_g = (c_g, \alpha_g, \beta_g, \ell_g)$ after obtaining each new observation. 
The exception is the conjoint correlation parameter $\rho_t$, which is marginalized by
sampling 50 linearly uniformly spaced points in $[-1, 1]$. Details in the Appendix.

%% file: content/experiments.tex

\section{Discussion and conclusion}

\begin{figure}
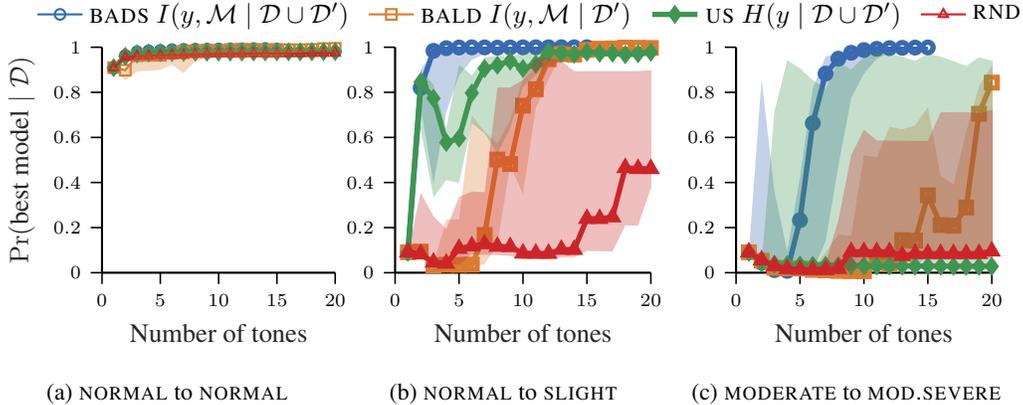

\begin{minipage}[b]{.32\linewidth}
\input{figures/normal_normal2}
\vspace{-2mm}
\subcaption{\acro{Normal} to \acro{Normal}}
\label{fig:performance_airline}
\end{minipage}%
\begin{minipage}[b]{.32\linewidth}
\centering
\input{figures/normal_slight2}
\vspace{-2mm}
\subcaption{\acro{Normal} to \acro{Slight}}
\end{minipage}
\begin{minipage}[b]{.32\linewidth}
\input{figures/moderate_modSevere2}
\vspace{-2mm}
\subcaption{\small{\acro{Moderate} to \acro{Mod.Severe}}}
\end{minipage}
\caption{Posterior probability of best model as function of number of iterations (tones delivered).
Solid lines represent the median. Upper and lower quartiles displayed by the shaded regions}
\label{fig:performance}
\vspace{-5mm}
\end{figure}

We want to evaluate if our approach, Bayesian Active Differential Selection (\acro{BADS}),
can quickly detect whether the patient's hearing thresholds are similar or different than a given reference exam. 
%

\textbf{\acro{NIOSH} database median audiogram generation}. To construct high-fidelity ground truth models, we used data from over 1 million audiograms available in the \acro{NIOSH} database \cite{masterson2013prevalence}. Each entry of this database includes 7 threshold intensity values per ear at 500 Hz, 1000 Hz, 2000 Hz, 3000 Hz, 4000 Hz, 6000 Hz and 8000 Hz. We classified each ear into one of seven categories of hearing loss, based on the pure-tone average (\acro{PTA}) of each ear, calculated by taking the mean of the threshold values at 500 Hz, 1000 Hz, and 2000 Hz. These categories are indicated in Table \ref{table:classification}A. Within each category, we generated a canonical audiogram by taking the median threshold value at each frequency. Ground truths were then extrapolated from these threshold values using a cubic spline interpolation in the frequency domain, followed by creating a sigmoid in the intensity domain. The resulting ground truth audiograms are presented in Figure \ref{fig:ground_truth}A.

\begin{table}[]
\centering
\caption{Number of iterations to achieve Bayes Factor equal or above 100 using \acro{BADS}. For each pair of (\acro{OLD}, \acro{NEW}) conditions, we performed
10 experiments. Average and standard deviation are shown. }
\smallskip\noindent
\resizebox{\linewidth}{!}{%
\begin{tabular}{lccccccc}
\toprule
\diagbox[width=7em]{\small{\acro{OLD}}}{\small{\acro{NEW}}}
   &  Normal  & Slight & Mild & Moderate & M. Severe & Severe & Profound \\
\midrule
Normal & 7.2 $\pm$ 1.6   & 5.8 $\pm$ 2.9 &    3.8 $\pm$ 0.79    & 3.7 $\pm$ 0.48    & 3.5 $\pm$ 0.53    & 3.7 $\pm$ 0.48    & 3.5 $\pm$ 0.53 \\
Slight & 5.0 $\pm$ 0.47 & 5.4 $\pm$ 0.70 & 9.9 $\pm$ 1.8 & 3.2 $\pm$ 0.42 & 3.3 $\pm$ 0.48 & 3.1 $\pm$ 0.32 & 3.0 $\pm$ 0 \\
Mild & 4.5 $\pm$ 0.53 & 6.7 $\pm$ 3.6 & 4.1 $\pm$ 0.32 & 12 $\pm$ 2.5 & 2.7 $\pm$ 0.48 & 2.9 $\pm$ 0.32 & 2.9 $\pm$ 0.32 \\
Moderate & 4.1 $\pm$ 0.32 & 4.5 $\pm$ 0.53 & 7.4 $\pm$ 4.1 & 3.8 $\pm$ 0.63 & 11 $\pm$ 2.6 & 3.0 $\pm$ 0 & 2.6 $\pm$ 0.52 \\
M. Severe & 4.4 $\pm$ 0.52 & 4.3 $\pm$ 0.48 & 7.5 $\pm$ 2.8 & 16 $\pm$ 8.0 & 3.9 $\pm$ 0.88 & 11 $\pm$ 2.6 & 2.8 $\pm$ 0.42 \\
Severe & 4.2 $\pm$ 0.63 & 4.2 $\pm$ 0.42 & 4.3 $\pm$ 0.48 & 8.2 $\pm$ 3.4 & 14 $\pm$ 9.2 & 4.2 $\pm$ 1.1 & 2.7 $\pm$ 0.48 \\
Profound & 3.0 $\pm$ 0 & 3.0 $\pm$ 0 & 3.5 $\pm$ 0.53 & 3.6 $\pm$ 0.84 & 3.3 $\pm$ 0.48 & 4.2 $\pm$ 0.63 & --- \\
\bottomrule
\end{tabular}
}
\vspace{-5mm}
\label{table:bf}
\end{table}

\textbf{Experimental setup}. These ground truths were then used to simulate patient's response of a given category or clinical condition.
We use the \acro{GP AMLAG} framework \cite{Song:2017, DiLorenzo2017ConjointAE} for obtaining 50 
initial observations $\data$, corresponding to the first exam. Specifically, we sample 15 points using the pseudo-random Halton sampling method and choose another 35 using the mutual information criterion between the output variable and the latent function $I(y^\ast; f)$ \cite{gardner2015psychophysical, houlsby2011bayesian}. This gives the initial data from the previous exam, or simply, \acro{OLD} exam. The new observations, representing the current or \acro{NEW} exam $\data'$, were selected by the methods compared below using the respective ground truth data for responses. All combinations of \acro{OLD} and \acro{NEW} exams result in 49 experimental conditions, and we further repeat each experiment 10 times. 


\textbf{Preliminary results}. First, we evaluate how quickly \acro{BADS} selects observations to differentiate between both models, $\model_f$ (same) and $\model_g$ (different). We compute how many iterations (tones) are needed to achieve a Bayes Factor \acro{BF} between these models equal to or greater than 100, suggesting strong evidence in favor of one of them. Because we know the ground truth model, we present the results in Table \ref{table:bf}, considering the \acro{BF} ratio in favor of the model that we consider to be ``correct''\footnote{We are not claiming correctness in its technical sense, but merely regarding this constructed experiment.}. The results indicate that, on average, fewer than 6 tones are required for differentiating between same or different models. Differentiating between more similar classes is more challenging, but even in these cases fewer than 20 tones are needed. The profound/profound case is an exception due to its degenerate nature (all responses are negative). In Table \ref{table:bads_results}A, we show the median, lower and upper quartiles of the posterior probability of the ``correct'' model as we obtain more observations. 

We also compare \acro{BADS} against three different active learning strategies. We kept the evaluation procedure the same, 
\textit{i.e.}, we compute the model posterior over both models considering all gathered data. 
Our first baseline is similar to how we collected the data from the previous exam, using
Bayesian active learning by disagreement \acro{BALD} that computes $I(y^\ast; f)$. This simulates a procedure that ignores 
previous information; hence during the decision-making part we simply compute the strategy with respect to the most recent data, 
$I(y^\ast; f \given \data')$. Our second baseline is an adaptation of uncertainty sampling (\acro{US}) considering both models and all data. 
The last baseline is random sampling (\acro{RND}). Figure \ref{fig:performance} shows results for three selected experiments. \acro{BADS} results were averaged
across 10 experiments during 15 iterations, and the baselines ran for more iterations and were averaged across 5 experiments. These initial
results indicate that \acro{BADS} outperforms the baselines. 

\textbf{Conclusion}. In this paper we proposed a novel framework for quickly determining whether a complex physiological system is in a different state than a reference. Previously, complete models of both systems would need to be estimated before their similarity could be evaluated. Bayesian active differential selection enables queries of the new system that are most informative in order to answer the question of whether it is the same as or different from the reference system. Efficiency gains are substantial, enabling accurate audiogram classification in a relatively small number of query tones.

%% file: content/appendix.tex
\newpage
\section{Appendix}

\textbf{Model  $\model_f$ predictive distributions}. The \acro{EP} algorithm renders a Gaussian approximation to the latent predictive distribution, $p( f^\ast \given \vec{x}^\ast, \data \cup \data', \model_f) \approx \mc{N}( \mu_{f \given \data}, \sigma^2_{f \given \data})$, which in turn
lets us compute the predictive distribution:
\begin{equation}
	p(y^\ast \given \vec{x}^\ast, \data \cup \data', \model_f) 
	= \mc{B} \Bigg(
	\Phi\bigg(  
		\frac{\mu_{f \given \data}(\vec{x}^\ast; \hat{\theta} )} 
		{\sqrt{1+\sigma^2_{f \given \data}(\vec{x}^\ast; \hat{\theta} )}}
	\bigg)
	\Bigg),
\end{equation}
which is a Bernoulli distribution. Let $p$ be the success probability, $ p = \Pr(y^\ast = 1 \given \vec{x}, \data \cup \data', \model_f)$, the differential entropy of this distribution is given by the Bernoulli entropy function $h$:
	$H[\mc{B}(p)] = h(p) = -p \log p - (1 - p) \log(1-p).$

\textbf{Model  $\model_g$ predictive distributions}. The conjoint correlation parameter $\rho_t$ is marginalized by sampling 50 linearly uniformly spaced points in $[-1, 1]$
This requires a different strategy for computing the predictive distributions: 

\begin{equation}
	\label{predictive_model_g}
	p(y^\ast \given \vec{x}^\ast, \data \cup \data', \model_g) 
	= \sum_j  
	\mc{B} 
	\Bigg(
	\Phi\bigg(
		\pi\frac{
				\mu_{g \given \data}(\vec{x}^\ast; \hat{\theta}_{g, j} )} 
			{
				\sqrt{1+\sigma^2_{g \given \data}(\vec{x}^\ast; \hat{\theta}_{g, j})}
			}
		\bigg)
	\Bigg)
	p(\rho_{t, j}),
\end{equation}
which is a uniform mixture of Bernoullis. Since Bernoulli distributions are closed under taking mixtures, we can readily compute the entropy of \eqref{predictive_model_g}. 

\textbf{Evidence and model-marginal predictive distribution computations}. For both models, we approximate the latent predictive distribution by a Gaussian distribution using the \acro{EP} algorithm. We have also assumed that the $\acro{MAP}$ point estimate for the hyperparameters is relatively reasonable for all but the task hyperparameter $\sigma^2_{t}$.
Using Bayes rule, we compute the posterior probability of each model given the data:
\begin{equation}
  \label{model_posterior}
  p(\model \given \data)
  = \frac{p(\vec{y} \given \mat{X}', \model)p(\model)}
          {p(\vec{y} \given \mat{X}')}
\end{equation}
using a simple quadrature rule for computing $p(\vec{y} \given \mat{X}', \model_g)$. For computing $p(y^\ast \given \vec{x}^\ast, \data \cup \data')$, we also note that this is a mixture of Bernoullis, and thus computing its entropy is straightforward. 

\newpage


\begin{figure}
\begin{minipage}[b]{.98\linewidth}
\centering
\input{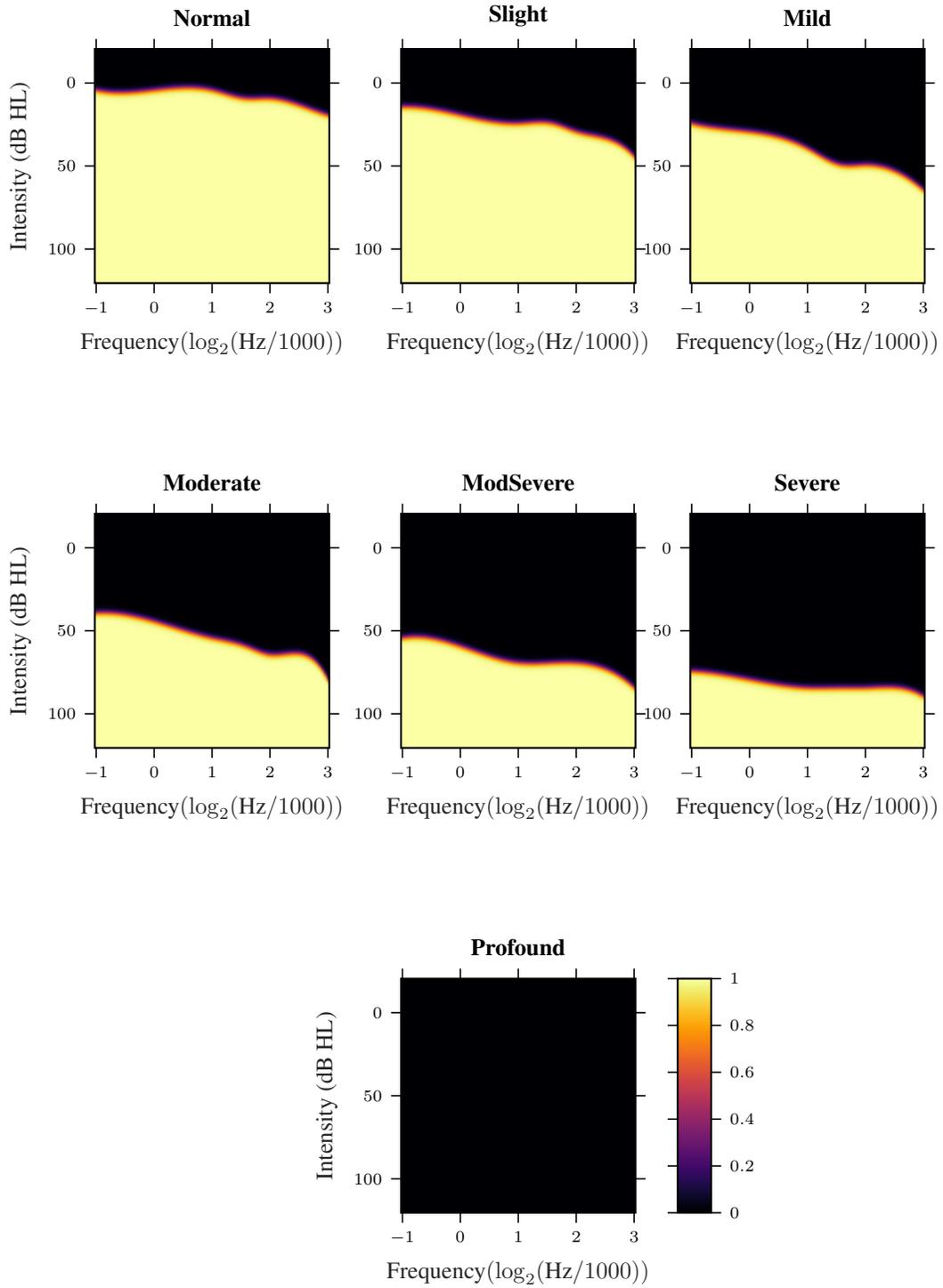}
\end{minipage}%
\caption{Ground truth audiograms of each hearing loss class.}
\label{fig:ground_truth}
\end{figure}


%

\begin{table}[]
\begin{center}
\begin{tabular}{ll}
\toprule
\textbf{Degree of hearing loss} & \textbf{Hearing loss range (dB HL)} \\ 
\midrule
Normal                          & -10 to 15                           \\
Slight                          & 16 to 25                            \\
Mild                            & 26 to 40                            \\
Moderate                        & 41 to 55                            \\
Moderately severe               & 56 to 70                            \\
Severe                          & 71 to 90                            \\
Profound                        & 91                                  \\ 
\bottomrule
\end{tabular}
\vspace{2mm}
\caption{Hearing loss classification using pure-tone average \cite{clark1981uses}.}
\label{table:classification}
\end{center}
\end{table}

\begin{table}[]
\smallskip\noindent
\resizebox{\linewidth}{!}{%
\begin{tabular}{lcccccccc}
& \multicolumn{7}{c}{\phantom{ppppppppppppppppppppp}\textbf{New conditions}} \\ 
\\
& & \acro{Normal} & \acro{Slight} & \acro{Mild} & \acro{Moderate} & \acro{M. Severe} & \acro{Severe} & \acro{Profound} \\
\multirow{7}{*}{\begin{turn}{90}\textbf{Old conditions}\phantom{pppppppppppppppppppppppppppppppppppppppppppp} \end{turn}}  & \begin{turn}{90}\phantom{space}\acro{Normal}\end{turn}    &  \input{figures/plot_bads_1} &  \input{figures/plot_bads_2} &  \input{figures/plot_bads_3} &  \input{figures/plot_bads_4} &  \input{figures/plot_bads_5} &  \input{figures/plot_bads_6} &  \input{figures/plot_bads_7}   \\
 \\
& \begin{turn}{90}\phantom{space} \acro{Slight}\end{turn}    &    \input{figures/plot_bads_8} &  \input{figures/plot_bads_9} &  \input{figures/plot_bads_10} &  \input{figures/plot_bads_11} &  \input{figures/plot_bads_12} &  \input{figures/plot_bads_13} &  \input{figures/plot_bads_14} \\
\\
& \begin{turn}{90}\phantom{spaceaa}  \acro{Mild}\end{turn}      &   \input{figures/plot_bads_15} &  \input{figures/plot_bads_16} &  \input{figures/plot_bads_17} &  \input{figures/plot_bads_18} &  \input{figures/plot_bads_19} &  \input{figures/plot_bads_20} &  \input{figures/plot_bads_21} \\
\\
& \begin{turn}{90}\phantom{spa}  \acro{Moderate}\end{turn}  &     \input{figures/plot_bads_22} &  \input{figures/plot_bads_23} &  \input{figures/plot_bads_24} &  \input{figures/plot_bads_25} &  \input{figures/plot_bads_26} &  \input{figures/plot_bads_27} &  \input{figures/plot_bads_28} \\
\\
& \begin{turn}{90}\phantom{spa}  \acro{M. Severe}\end{turn} &     \input{figures/plot_bads_29} &  \input{figures/plot_bads_30} &  \input{figures/plot_bads_31} &  \input{figures/plot_bads_32} &  \input{figures/plot_bads_33} &  \input{figures/plot_bads_34} &  \input{figures/plot_bads_35} \\
\\
& \begin{turn}{90}\phantom{space}  \acro{Severe}\end{turn}    &       \input{figures/plot_bads_36} &  \input{figures/plot_bads_37} &  \input{figures/plot_bads_38} &  \input{figures/plot_bads_39} &  \input{figures/plot_bads_40} &  \input{figures/plot_bads_41} &  \input{figures/plot_bads_42} \\
\\
& \begin{turn}{90}\phantom{spac}  \acro{Profound}\end{turn}  &   \input{figures/plot_bads_43} &  \input{figures/plot_bads_44} &  \input{figures/plot_bads_45} &  \input{figures/plot_bads_46} &  \input{figures/plot_bads_47} &  \input{figures/plot_bads_48} &  \input{figures/plot_bads_49} \\
\end{tabular}
}
\caption{Median probability of the best model for \acro{BADS} across 10 repetitions for all 49 combinations of (\acro{OLD}, \acro{NEW}) new experiments. Upper and lower quartiles displayed by the shaded regions.}
\label{table:bads_results}
\end{table}

%% file: neurips_2019.bbl
\begin{thebibliography}{21}
\providecommand{\natexlab}[1]{#1}
\providecommand{\url}[1]{\texttt{#1}}
\expandafter\ifx\csname urlstyle\endcsname\relax
  \providecommand{\doi}[1]{doi: #1}\else
  \providecommand{\doi}{doi: \begingroup \urlstyle{rm}\Url}\fi

\bibitem[Barbour et~al.(2019)Barbour, DiLorenzo, Sukesan, Song, Chen, Degen,
  Heisey, and Garnett]{Barbour2019}
Dennis~L. Barbour, James~C. DiLorenzo, Kiron~A. Sukesan, Xinyu~D. Song, Jeff~Y.
  Chen, Eleanor~A. Degen, Katherine~L. Heisey, and Roman Garnett.
\newblock Conjoint psychometric field estimation for bilateral audiometry.
\newblock \emph{Behavior Research Methods}, 51\penalty0 (3):\penalty0
  1271--1285, 2019.

\bibitem[Carhart and Jerger(1959)]{carhart1959preferred}
Raymond Carhart and James~F Jerger.
\newblock Preferred method for clinical determination of pure-tone thresholds.
\newblock \emph{Journal of speech and hearing disorders}, 24\penalty0
  (4):\penalty0 330--345, 1959.

\bibitem[Cavagnaro et~al.(2010)Cavagnaro, Myung, Pitt, and
  Kujala]{cavagnaro2010adaptive}
Daniel~R Cavagnaro, Jay~I Myung, Mark~A Pitt, and Janne~V Kujala.
\newblock Adaptive design optimization: A mutual information-based approach to
  model discrimination in cognitive science.
\newblock \emph{Neural computation}, 22\penalty0 (4):\penalty0 887--905, 2010.

\bibitem[Clark et~al.(1981)]{clark1981uses}
John~G Clark et~al.
\newblock Uses and abuses of hearing loss classification.
\newblock \emph{Asha}, 23\penalty0 (7):\penalty0 493--500, 1981.

\bibitem[DiLorenzo(2017)]{DiLorenzo2017ConjointAE}
James~C DiLorenzo.
\newblock Conjoint audiogram estimation via gaussian process classification.
\newblock In \emph{Master of Science thesis}, 2017.

\bibitem[Gardner et~al.(2015{\natexlab{a}})Gardner, Malkomes, Garnett,
  Weinberger, Barbour, and Cunningham]{Gardner:2015}
Jacob Gardner, Gustavo Malkomes, Roman Garnett, Kilian~Q Weinberger, Dennis
  Barbour, and John~P Cunningham.
\newblock Bayesian active model selection with an application to automated
  audiometry.
\newblock In \emph{Advances in Neural Information Processing Systems 28}, pages
  2386--2394. 2015{\natexlab{a}}.

\bibitem[Gardner et~al.(2015{\natexlab{b}})Gardner, Song, Weinberger, Barbour,
  and Cunningham]{gardner2015psychophysical}
Jacob~R. Gardner, Xinyu Song, Kilian~Q. Weinberger, Dennis Barbour, and John~P.
  Cunningham.
\newblock {Psychophysical Detection Testing with Bayesian Active Learning}.
\newblock In \emph{UAI}, 2015{\natexlab{b}}.

\bibitem[Garnett et~al.(2014)Garnett, Osborne, and Hennig]{garnett2013active}
Roman Garnett, Michael~A Osborne, and Philipp Hennig.
\newblock {Active Learning of Linear Embeddings for Gaussian Processes}.
\newblock In \emph{UAI}, pages 230--239, 2014.

\bibitem[Hamburg and Collins(2010)]{Hamburg:2010}
Margaret~A. Hamburg and Francis~S. Collins.
\newblock The path to personalized medicine.
\newblock \emph{New England Journal of Medicine}, 363\penalty0 (4):\penalty0
  301--304, 2010.
\newblock PMID: 20551152.

\bibitem[Hern{\'a}ndez-Lobato et~al.(2014)Hern{\'a}ndez-Lobato, Hoffman, and
  Ghahramani]{hl2014pes}
Jos{\'e}~Miguel Hern{\'a}ndez-Lobato, Matthew~W Hoffman, and Zoubin Ghahramani.
\newblock {Predictive Entropy Search for Efficient Global Optimization of
  Black-box Functions}.
\newblock In \emph{NIPS}, pages 918--926, 2014.

\bibitem[Houlsby et~al.(2012)Houlsby, Huszar, Ghahramani, and
  Hern{\'a}ndez-Lobato]{houlsby2011bayesian}
Neil Houlsby, Ferenc Huszar, Zoubin Ghahramani, and Jose~M
  Hern{\'a}ndez-Lobato.
\newblock {Collaborative Gaussian Processes for Preference Learning}.
\newblock In \emph{NIPS}, pages 2096--2104, 2012.

\bibitem[Houlsby et~al.(2014)Houlsby, Hern\'{a}ndez-Lobato, and
  Ghahramani]{houlsby2014cold}
Neil Houlsby, Jos{\'e}~M Hern\'{a}ndez-Lobato, and Zoubin Ghahramani.
\newblock {Cold-start Active Learning with Robust Ordinal Matrix
  Factorization}.
\newblock In \emph{ICML}, pages 766--774, 2014.

\bibitem[Kononenko(2001)]{KONONENKO200189}
Igor Kononenko.
\newblock Machine learning for medical diagnosis: history, state of the art and
  perspective.
\newblock \emph{Artificial Intelligence in Medicine}, 23\penalty0 (1):\penalty0
  89 -- 109, 2001.
\newblock ISSN 0933-3657.
\newblock \doi{https://doi.org/10.1016/S0933-3657(01)00077-X}.
\newblock URL
  \url{http://www.sciencedirect.com/science/article/pii/S093336570100077X}.

\bibitem[Leek(2001)]{Leek2001}
Marjorie~R. Leek.
\newblock Adaptive procedures in psychophysical research.
\newblock \emph{Perception {\&} Psychophysics}, 63\penalty0 (8):\penalty0
  1279--1292, Nov 2001.
\newblock ISSN 1532-5962.
\newblock \doi{10.3758/BF03194543}.
\newblock URL \url{https://doi.org/10.3758/BF03194543}.

\bibitem[Masterson et~al.(2013)Masterson, Tak, Themann, Wall, Groenewold,
  Deddens, and Calvert]{masterson2013prevalence}
Elizabeth~A Masterson, SangWoo Tak, Christa~L Themann, David~K Wall, Matthew~R
  Groenewold, James~A Deddens, and Geoffrey~M Calvert.
\newblock Prevalence of hearing loss in the united states by industry.
\newblock \emph{American journal of industrial medicine}, 56\penalty0
  (6):\penalty0 670--681, 2013.

\bibitem[Minka(2001)]{minka2001expectation}
Thomas~P Minka.
\newblock Expectation propagation for approximate bayesian inference.
\newblock In \emph{Proceedings of the Seventeenth conference on Uncertainty in
  artificial intelligence}, pages 362--369. Morgan Kaufmann Publishers Inc.,
  2001.

\bibitem[Rasmussen and Williams(2006)]{RasWil06}
C.~E. Rasmussen and C.~K.~I. Williams.
\newblock \emph{{Gaussian Processes for Machine Learning}}.
\newblock MIT Press, 2006.

\bibitem[Schlittenlacher et~al.(2018)Schlittenlacher, Turner, and
  Moore]{Schlittenlacher:2018}
Josef Schlittenlacher, Richard~E. Turner, and Brian C.~J. Moore.
\newblock Audiogram estimation using bayesian active learning.
\newblock \emph{The Journal of the Acoustical Society of America}, 144\penalty0
  (1):\penalty0 421--430, 2018.

\bibitem[Shademan et~al.(2016)Shademan, Decker, Opfermann, Leonard, Krieger,
  and Kim]{Shademan337ra64}
Azad Shademan, Ryan~S. Decker, Justin~D. Opfermann, Simon Leonard, Axel
  Krieger, and Peter C.~W. Kim.
\newblock Supervised autonomous robotic soft tissue surgery.
\newblock \emph{Science Translational Medicine}, 8\penalty0 (337):\penalty0
  337ra64--337ra64, 2016.
\newblock ISSN 1946-6234.

\bibitem[Song et~al.(2015)Song, Wallace, Gardner, Ledbetter, Weinberger, and
  Barbour]{Song:2015}
X.~D. Song, B.~M. Wallace, J.~R. Gardner, N.~M. Ledbetter, K.~Q. Weinberger,
  and D.~L. Barbour.
\newblock {Fast, Continuous Audiogram Estimation Using Machine Learning.}
\newblock In \emph{Ear and hearing}, pages 326--335, 2015.

\bibitem[Song et~al.(2017)Song, Garnett, and Barbour]{Song:2017}
Xinyu~D. Song, Roman Garnett, and Dennis~L. Barbour.
\newblock Psychometric function estimation by probabilistic classification.
\newblock \emph{The Journal of the Acoustical Society of America}, 141\penalty0
  (4):\penalty0 2513--2525, 2017.
\newblock \doi{10.1121/1.4979594}.
\newblock URL \url{https://doi.org/10.1121/1.4979594}.

\end{thebibliography}
